# Physics-informed AI Accelerated Retention Analysis of Ferroelectric Vertical NAND: From Day-Scale TCAD to Second-Scale Surrogate Model

Gyujun Jeong[1], Sungwon Cho[1], Minji Shon[1], Namhoon Kim[1], Woohyun Hwang[2], Kwangyou Seo[2], Suhwan Lim[2], Wanki Kim[2], Daewon Ha[2], Prasanna Venkatesan[3], Kihang Youn[3], Ram Cherukuri[3], Yiyi Wang[3], Suman Datta[1], Asif Khan[1], and Shimeng Yu[1]

[1] School of Electrical and Computer Engineering, Georgia Institute of Technology, GA, USA,

[2] Semiconductor Research and Development, Samsung Electronics Co., Ltd, South Korea

[3] NVIDIA, Santa Clara, CA, USA

Email: gjeong35@gatech.edu, shimeng.yu@ece.gatech.edu

***Abstract*— Ferroelectric field-effect transistors (FeFET)-based vertical NAND (Fe-VNAND) has emerged as a promising candidate to overcome z-scaling limitations with lower programming voltages. However, the data retention of 3D Fe-VNAND is hindered by the complex interaction between charge detrapping and ferroelectric depolarization. Developing optimized device designs requires exploring an extensive parameter space, but the high computational cost of conventional Technology Computer-Aided Design (TCAD) tools makes such wide-scale optimization impractical. To overcome these simulation barriers, we present a Physics-Informed Neural Operator (PINO)-based AI surrogate model designed for high-efficiency prediction of threshold voltage (Vth) shifts and retention behavior. By embedding fundamental physical principles into the learning architecture, our PINO framework achieves a speedup exceeding 10000× compared to TCAD while maintaining physical accuracy. The resulting surrogate provides a physics-consistent data engine for compact model parameter extraction and look-up-table (LUT) generation, directly supporting reliability-aware SPICE simulation of Fe-VNAND. This study demonstrates the model's effectiveness on a single FeFET configuration, serving as a pathway toward modeling the retention loss mechanisms.**

## I. Introduction

As the demand for high-density storage continues to grow, 3D Vertical NAND (VNAND) flash memory has become essential. However, continued z-scaling faces significant bottlenecks, primarily due to the high programming voltages required by conventional charge trap layers [1]. Ferroelectric field-effect transistors (FeFETs) have emerged as a promising candidate, offering lower operating voltages and faster switching speeds. Despite these advantages, the reliability of Fe-VNAND, particularly data retention, poses a crucial challenge. The retention characteristics are governed by a complex, time-dependent interplay between ferroelectric polarization loss (depolarization) and charge trapping and detrapping dynamics at the dielectric interfaces [2]. Optimizing the gate stack to mitigate these effects requires precise engineering of the Tunnel Dielectric Layer (TDL) and ferroelectric geometry [3]. Analyzing these mechanisms typically relies on Technology Computer-Aided Design (TCAD) simulations. While accurate, TCAD is computationally prohibitive for large-scale design space exploration (DSE). A single rigorous simulation sweeping temperature and time can take over 24 hours. When optimizing across multiple variables—such as dimension, temperature, and channel geometry—the computational cost scales linearly, making comprehensive optimization infeasible. To bridge the gap between physical accuracy and computational efficiency, this work proposes a Physics-Informed AI Surrogate Model. Unlike standard black-box machine learning approaches, which often violate physical laws (e.g., non-monotonic charge detrapping), our model utilizes a Physics-Informed Neural Operator (PINO) architecture [4]. By enforcing governing equations via the loss function, the model predicts 2D physics maps and retention characteristics with near-TCAD accuracy but at speeds enabling real-time inference, thereby serving as a rapid data source for compact model parameter extraction and reliability-aware LUT generation in Fe-VNAND circuit simulation.

## II. Device Physics and TCAD Modeling Methodology

### A. Fe-VNAND Device Structure

This study utilizes a single FeFET device featuring an interlayer-engineered gate stack, specifically incorporating a Tunnel Dielectric Layer (TDL) within the ferroelectric layers (HZO/TDL/HZO), as illustrated in Fig. 1(a). Building on the foundation that the TDL functions as a charge-screening region [2], we analyze the erase (ERS) operation where a strong negative gate bias triggers ferroelectric polarization switching. Simultaneously, charges accumulate at the HZO/TDL interface via Fowler-Nordheim (FN) tunneling, effectively counteracting the depolarization field to stabilize the state and expand the Memory Window (MW). For the AI surrogate modeling and retention analysis, this work focuses on a more active transient regime by employing a modified physical parameterization. This configuration induces a pronounced, time-dependent degradation profile driven by the interplay of charge detrapping

This work was supported by Samsung Electronics (Award no. IO250304-12193-01).

and depolarization. Given that the proposed PINO framework is designed to learn the dynamic, time-varying partial differential equations (PDEs) governing device reliability, these complex transient characteristics provide a rigorous benchmark for validating the AI's predictive fidelity. While this work primarily details the ERS state for a focused physical analysis, preliminary evaluations indicate that the program (PGM) state exhibits symmetric behavior, making the proposed modeling approach equally applicable to both states.

### B. Physics of Retention Loss

Retention loss in this structure is driven by two mechanisms: ferroelectric depolarization and charge detrapping (Fig. 1(b)) [5]. The depolarization field is an internal electric field generated by, and opposing, the ferroelectric polarization state, which causes a gradual loss of remnant polarization ($P_r$) over time. Simultaneously, charge detrapping occurs from the dielectric interfaces. The loss of these screening charges gradually exacerbates the internal field, thereby accelerating the degradation of the ferroelectric polarization.

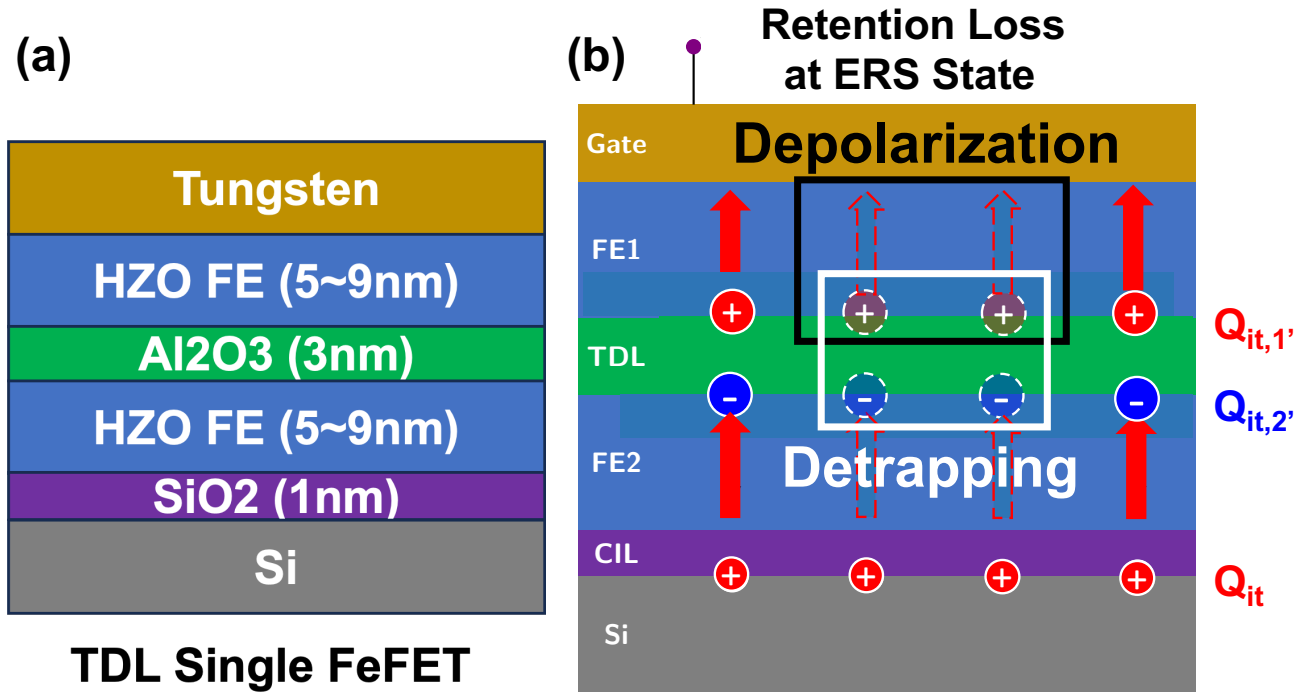

Fig. 1: (a) Device structure of a single FeFET with tunnel dielectric layer (TDL) and HZO FE layer. (b) Visualization of the charge detrapping and depolarization on FeFET retention in the erase state.

### C. TCAD Training Data Synthesis

To generate high-fidelity training data, we employed Synopsys Sentaurus TCAD. The simulation framework integrates the ferroelectric Preisach model and Fowler-Nordheim (FN) tunneling [2]. The ferroelectric Preisach model captures the hysteretic polarization switching and history dependence. Time-dependent charge trapping and detrapping are modeled to simulate retention up to $10^4$ seconds. The dataset was constructed by sweeping 3 input parameters in Table I: HZO thickness ($t_{HZO}$), ambient temperature ($T$), and retention time ($\tau$). For each condition, the TCAD solver outputs the $I_D$-$V_G$ characteristics and their corresponding internal 2D spatial physics maps. To maximize the surrogate model's resolution and explicitly target the mechanisms driving retention degradation, the spatial extraction of the electrostatic potential, polarization field, and trapped carrier densities (both electron and hole) was strictly restricted to the engineered gate-stacked region (HZO/TDL/HZO). Concurrently, the 2D current density map was extracted from entire device structure to capture the carrier transport behavior and channel conductance changes within the active silicon channel (Fig. 2).

TABLE I. INPUT SWEEP SPACE FOR TCAD DATA SYNTHESIS

| Input Sweep Parameters | Values |
|---|---|
| HZO Thickness | {5, 6, 7, 8, 9} nm |
| Temperature | {300, 358, 400, 473} K |
| Retention Time | {0, 1e0, 1e1, 1e2, 1e3, 1e4} s |

Fig. 2: Output intermediate physics maps of the AI surrogate model. The internal device states, including electrostatic potential, ferroelectric polarization, and trapped carrier densities (e-/h+), are extracted from the engineered gate-stack region (HZO/TDL/HZO) to resolve localized reliability mechanisms. In contrast, the current density (J) is extracted across the entire device domain to accurately capture the modulation of carrier transport within the silicon channel.

## III. PHYSICS-INFORMED AI ARCHITECTURE

The proposed surrogate model is implemented using the NVIDIA PhysicsNeMo™ framework [6]. It utilizes a multi-phase architecture designed to learn the solution operator of the underlying PDEs and physics laws.

### A. Architecture Overview

The pipeline proceeds through three distinct phases (Fig. 3). Phase 1 initiates the workflow by embedding the scalar device inputs—specifically $t_{HZO}$, $T$ and logarithmic retention time $\log(\tau)$. These embedded features are subsequently fed into Phase 2, which serves as the core physics surrogate engine. The backbone of this engine is a Fourier Neural Operator (FNO). In contrast to conventional Convolutional Neural Networks (CNNs) that process localized, discrete pixel data, FNOs perform global convolutions within the Fourier domain [4]. This architectural advantage allows the network to learn the continuous solution operator of the governing PDEs, thereby mapping the scalar inputs to resolution-invariant 2D physical fields in a single inference step. Crucially, generating these intermediate physical maps—rather than employing a black-box model to directly map scalar inputs (e.g., HZO thickness, temperature, retention time) to $I_D$-$V_G$ curves—is essential for establishing a physically interpretable digital twin. It provides physical explainability by allowing engineers to verify localized charge detrapping dynamics, and it acts as a physical regularizer, ensuring the downstream readout network only processes physically valid internal device states. To transition this purely data-driven FNO into a PINO, governing physics-informed loss functions are embedded into the training; the specific mathematical formulation of these physical constraints is detailed in the subsequent section. To optimize memory utilization and balance computational efficiency, the FNO training is evaluated on a 256×256 spatial grid. Phase 3 utilizes an IV-Net, a hybrid architecture combining a CNN and a Multilayer Perceptron (MLP), to predict final device electrical metrics such as $I_D$-$V_G$ curves. To achieve this, we employ a scalar context injection. A CNN-based map encoder extracts spatial features from the Phase 2 physics maps. Simultaneously, the original scalar inputs (from Phase 1) bypass the physics surrogate and are injected directly into a parallel MLP. The outputs are then fused via weighted summation. This residual connection is essential because the global thermodynamic context, introduced by elevated temperature during retention baking, cannot be extracted from purely electrostatic 2D maps. By injecting these scalars, the model preserves global context and corrects for quantization errors, significantly improving prediction accuracy. To prevent overfitting on the highly constrained training dataset of 120 TCAD sweeps, the network architecture was strictly regularized. Deep learning models with millions of parameters would perfectly memorize this sparse dataset and fail on unseen data. Therefore, the FNO utilized a low-frequency mode cutoff ($modes = 12$) across four Fourier layers with narrow latent channels ($W = 32$) to avoid capturing high-frequency noise. Furthermore, the readout IV-Net avoided deep standard architectures (e.g., VGG or ResNet). Instead, it employed a shallow CNN, reducing the parameter volume.

## *B. Physics-Informed Loss Function*

A fundamental distinction between the standard, purely data-driven FNO and physics-informed PINO is the implementation of a dual-loss formulation during the training phase. The overall objective function is defined as follows:

$$\mathcal{L}_{total} = \mathcal{L}_{data} + \lambda \cdot \mathcal{L}_{physics} \tag{1}$$

The data loss term ($L_{data}$) computes the L2 norm error between the neural network's predicted 2D physical fields or $I_D$-$V_G$ curves and the TCAD-generated ground truth. To transition the model from a black-box approximator to a PINO, the physics loss ($L_{physics}$) is introduced as a regularizer. This term enforces physical consistency, which is necessary to constrain the hypothesis space in the absence of dense training data. We employed two primary $L_{physics}$ components: Poisson residual and monotonicity constraints.

**1. Poisson Residual**

This term penalizes deviations from the governing electrostatics. It ensures that the predicted electrostatic potential ($\phi$), charge density ($\rho$), and ferroelectric polarization (P) follow Poisson's equation:

$$\mathcal{L}_{Poisson} = \nabla \cdot (\epsilon \nabla \phi) + (\rho - \nabla \cdot \mathrm{P}) \tag{2}$$

**2. Monotonicity constraints**

As implemented in [7], this constraint enforces the monotonic decay of both the remnant polarization and the trapped charge over the retention time:

$$\mathcal{L}_{mono} = ReLU\left(\frac{\partial |P_r|}{\partial \tau}\right) + ReLU\left(\frac{\partial |Q_{it}|}{\partial \tau}\right) \tag{3}$$

By penalizing unphysical transient glitches or temporal oscillations, this constraint ensures realistic depolarization and charge detrapping dynamics, thereby significantly enhancing the model's physical fidelity and generalization performance across unseen temporal and thermal regimes.

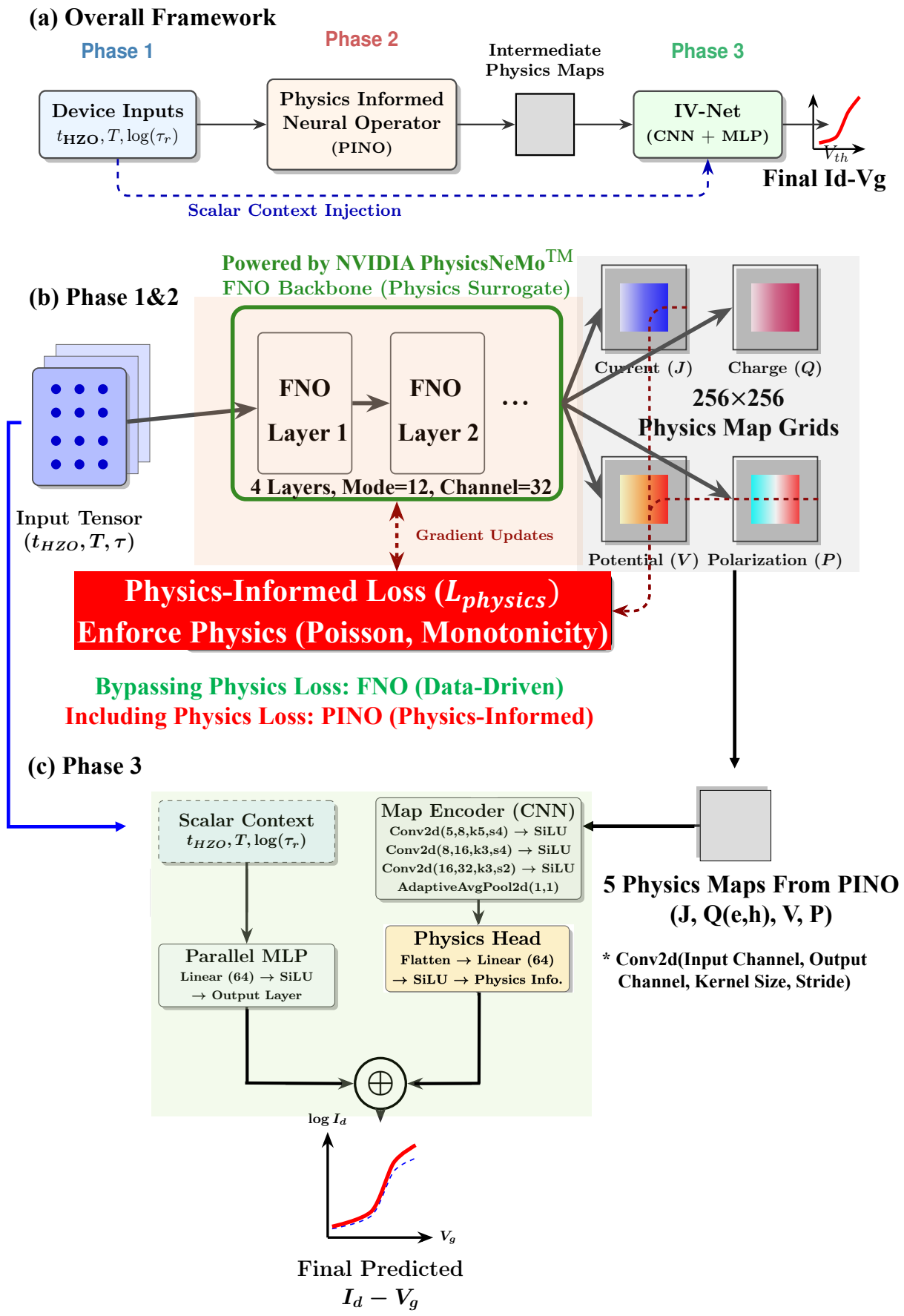

Fig. 3: AI Surrogate Model Framework built on NVIDIA PhysicsNeMo™. (a) Overall Framework. (b) Phase 1&2: Input Parameter Embedding and PINO/FNO surrogate for 2D physics fields prediction; inclusion of physics loss (red box) enables physics-informed PINO operation, while bypassing it results in a data-driven FNO. (d) Phase 3: Device $I_D$-$V_G$ readout (IV-Net) comprised of CNN and MLP to predict $I_D$-$V_G$ curve from given input condition and physics map from PINO.

# IV. Results and Discussion

## *A. Reconstruction Accuracy*

The trained PINO model demonstrates high fidelity in reconstructing internal device states. When validated against TCAD datasets, the model predicts 2D profiles of electrostatic potential, trapped charge, and current density with a relative error of ~ 0.1% (Fig. 4).

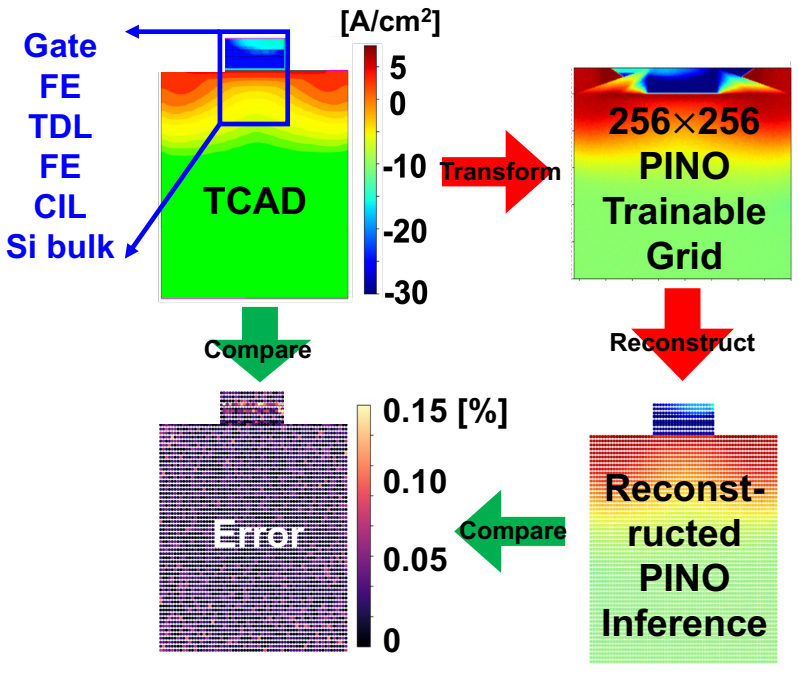

Fig. 4: Validation on Single and FeFET 2D Physics (current density) maps (a) TCAD Ground Truth. (b) PINO-trainable 256×256 Grid. (c) Reconstructed AI Inference. (d) Error map showing error ~ 0.1% comparing TCAD and reconstructed PINO inference.

## B. $I_D$-$V_G$ Prediction and Retention Analysis

Fig. 5 illustrates the retention prediction capabilities. The IV-Net backend converts these physics maps into electrical characteristics. The model achieves an $R^2 > 0.998$ for threshold voltage (Vth) prediction, measured at $I_D$=100nA, over the entire parameter sweep. The model accurately interpolates the Vth shift caused by charge detrapping and depolarization over $10^4$s, matching TCAD projections. By leveraging the physics-informed constraint, the model also demonstrates physically consistent interpolation for unseen temperatures (e.g., predicting 350K behavior between seen 300K and 400K). Crucially, the physics-informed loss successfully eliminates unphysical artifacts. As shown in the $I_D$-$V_G$ analysis (Fig. 6), purely data-driven models are prone to overfitting the sparse dataset, exhibiting unstable ripples across the transfer curve. In contrast, the PINO model, constrained by physics loss, correctly predicts a smooth $I_D$-$V_G$ response. The PINO model achieves superior generalization, improving the overall root-mean-squared-error (RMSE) by > 40% compared to the purely data-driven baseline.

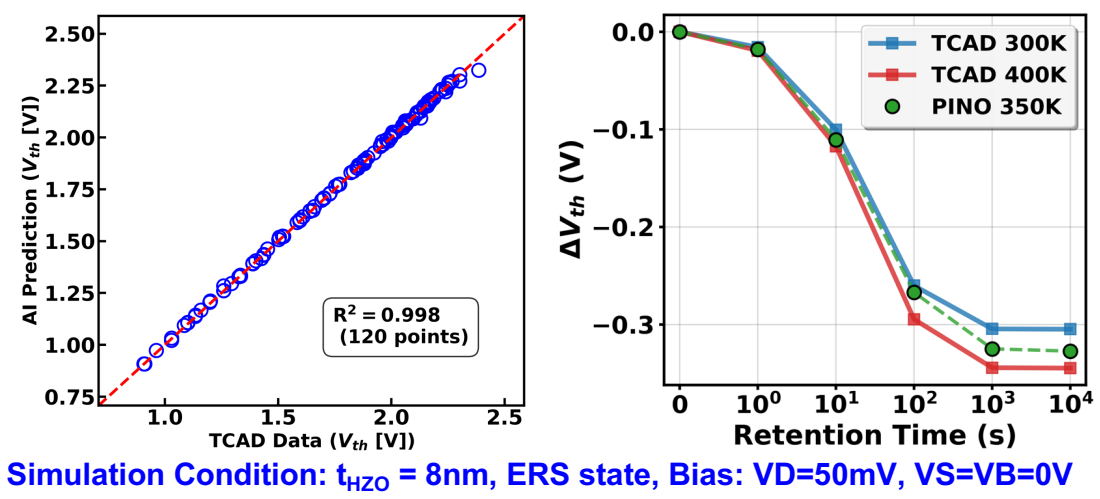

Fig. 5: Performance validation of the AI surrogate model. (a) Accuracy validation of predicted $Vth$ against TCAD ground truth across the full device parameter sweep, achieving an $R^2$ = 0.998 over 120 seen datapoints. (b) Retention loss ($\Delta Vth$) interpolation capability for an unseen temperature (T = 350K), demonstrating physically consistent retention loss tracking between 300K and 400K TCAD baselines.

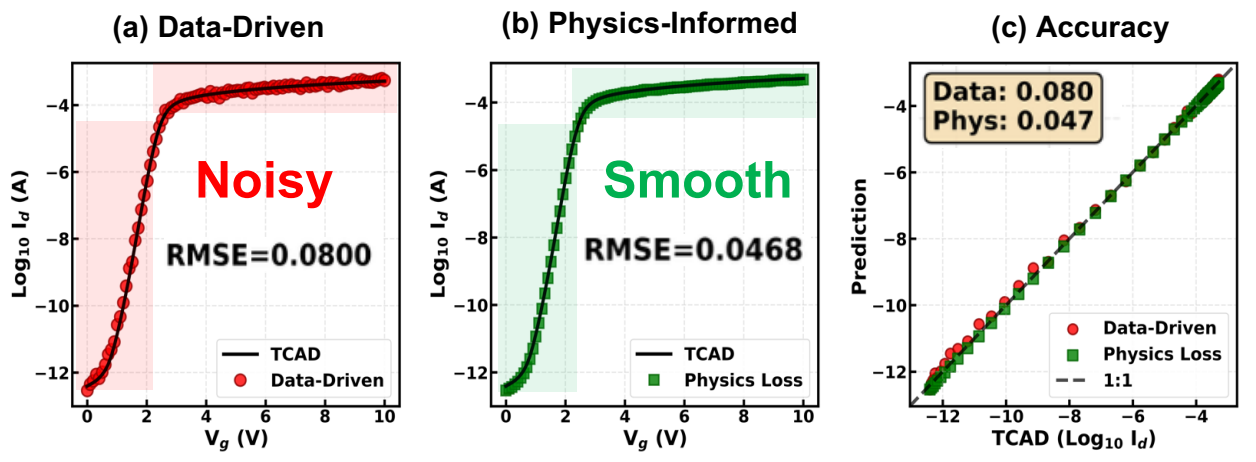

Fig. 6: Data-driven vs. Physics-Informed Comparison. (a) Data-driven: $I_D$-$V_G$ curve exhibits unstable ripples and lower accuracy. (b) Physics-informed: $I_D$-$V_G$ curve aligns closely with TCAD ground truth. (c) Accuracy: RMSE improved by > 40% with physics-informed loss.

## C. Computational Efficiency and Design Space Exploration

The primary advantage of the proposed AI surrogate is its computational efficiency. Table II benchmarks the computational cost of the PINO framework against conventional TCAD. The surrogate accelerates the inference time from 60 hours in TCAD to approximately 10 seconds for a full retention sweep, a speedup of four orders of magnitude.

TABLE II. COMPUTATIONAL EFFICIENCY BENCHMARK*

| Metric | TCAD | AI Surrogate |
| --- | --- | --- |
| **Single sweep** | 30 min | ~ 0.1s (real-time) |
| **Full sweep** | 60 hrs | ~ 10s |
| **DSE coverage** | Discrete (Prone to blind spots) | Continuous (mesh-free interpolation) |
| **Cost of optimization** | Prohibitive | Minimal |

* Benchmarks performed on Intel Xeon Platinum 8470N, 52-core CPU (TCAD) and NVIDIA Titan RTX 24 GB GPU (AI Surrogate).

# V. CONCLUSION

We demonstrated a physics-informed AI surrogate model that accelerates Fe-VNAND retention analysis by orders of magnitude. By validating the PINO model for single FeFET retention dynamics by embedding physical laws into the loss function, the framework successfully generalizes to unseen temporal, thermal, and geometric regimes, overcoming the data sparsity inherent in expensive semiconductor simulations. This work establishes a foundation for full-scale 3D Fe-VNAND retention analysis and the downstream development of reliability-aware compact models capturing coupled retention dynamics for SPICE-level simulation.

# ACKNOWLEDGMENT

This work was supported by Samsung Electronics (IO250304-12193-01).

# REFERENCES

[1] S. Yoon et al., "Highly stackable 3D ferroelectric NAND devices: Beyond the charge trap based memory," *Proc. IEEE IMW*, 2022.

[2] M. Shon et al., "A Comprehensive Modeling of Gate Stack Interlayer Engineering for Ferroelectric Vertical NAND," *Proc. IEEE DRC*, 2025.

[3] P. Venkatesan et al., "Pushing the limits of NAND technology scaling with ferroelectrics,". MRS Bulletin 50, 2025. https://doi.org/10.1557/s43577-025-00991-y.

[4] Z. Li et al., "Fourier Neural Operator for Parametric Partial Differential Equations," *arXiv:2010.08895*, 2020.

[5] P. Venkatesan et al., "Disturb and its mitigation in ferroelectric field-effect transistors with large memory window for NAND flash applications,". IEEE EDL. 45(12), 2367, 2024.

[6] NVIDIA PhysicsNeMo Framework, https://developer.nvidia.com/physicsnemo.

[7] G. Jeong et al., "Physics-Informed Neural Network-Assisted Compact Modeling for Process Variation in Non-Volatile Capacitive Crossbar Arrays," *Proc. IEEE IRPS, 2026.*